\documentclass{elsarticle}

\usepackage{graphicx}      % include this line if your document contains figures
\usepackage{amsmath, amssymb}
\usepackage{epstopdf}
\usepackage{array}
\usepackage{float}
\usepackage{color}
\usepackage{blindtext}
\usepackage{algorithmicx}
\usepackage[ruled]{algorithm}
\usepackage{algpseudocode}
\usepackage{algpascal}
\usepackage{algc}

\newcommand{\Real}{\mathbb{R}}
\newcommand{\Natural}{\mathbb{N}}

\newcommand{\Cov}{\mathrm{Cov}}

\newtheorem{rem}{Remark}

\newtheorem{theorem}{Theorem}

\usepackage{tikz}

\usetikzlibrary{arrows,shapes,positioning,shadows,trees}
\tikzstyle{decision} = [diamond, draw, text width=4.5em, text badly centered, node distance=1.5cm, inner sep=0pt]
\tikzstyle{block} = [rectangle, draw, text width=5em, text centered, rounded corners, minimum height=4em]

\begin{document}

\begin{frontmatter}

\title{Knowledge Transfer Between Artificial Intelligence Systems}

\author[Leic,Leti,UnivAI]{Ivan Yu. Tyukin\corref{cor1}}\ead{I.Tyukin@le.ac.uk} %\fnref{f1}
\author[Leic,UnivAI]{Alexander N. Gorban}\ead{ag153@le.ac.uk}
\author[Leic,ARM]{Konstantin I. Sofeikov}\ead{sofeykov@gmail.com}
\author[ARM]{Ilya Romanenko}\ead{Ilya.Romanenko@arm.com}

\address[Leic]{University of Leicester, Department of Mathematics, University Road, Leicester, LE1 7RH, UK}
\address[Leti]{Department of Automation and
        Control Processes, St. Petersburg State University of
        Electrical Engineering, Prof. Popova str. 5, Saint-Petersburg, 197376, Russian Federation}
\address[UnivAI]{UnivAI Ltd, 5 Park Court, Pyrford Road, West Byfleet, KT14 6SD, UK}
\address[ARM]{Imaging and Vision Group, ARM Ltd, 1 Summerpool Rd, Loughborough, LE11 5RD, UK}

\cortext[cor1]{Corresponding author}
%\fntext[f1]{The work was supported by Innovate UK Technology Strategy Board (Knowledge Transfer Partnership grants KTP009890 and KTP010522).}

\begin{abstract}
We consider the fundamental question: how a legacy ``student'' Artificial Intelligent (AI) system could learn from a legacy ``teacher'' AI system or a human expert without complete re-training and, most importantly, without requiring significant computational resources. Here ``learning'' is understood as an ability of one system to mimic responses of the other and vice-versa. We call such learning an Artificial Intelligence knowledge transfer. We show that if internal variables of the  ``student'' Artificial Intelligent system have the structure of an $n$-dimensional topological vector space and $n$ is sufficiently high then, with probability close to one, the required knowledge transfer can be implemented by simple cascades of linear functionals. In particular, for $n$ sufficiently large, with probability close to one, the ``student'' system can successfully and non-iteratively learn $k\ll n$ new examples from the ``teacher'' (or correct the same number of mistakes) at the cost of two additional inner products. The concept is illustrated with an example of knowledge transfer from a pre-trained convolutional neural network to a simple linear classifier with HOG features.
\end{abstract}


\end{frontmatter}

\section{Introduction}

Knowledge transfer between Artificial Intelligent systems has been the subject of extensive discussion in the literature for more than two decades \cite{Gorban:DAN:1991}, \cite{Hinton:NC:1991}, \cite{Pratt:ANIP:1992}, \cite{Schultz:2000}. State-of-the art approach to date is to use, or salvage, parts of the “teacher” AI system in the “student” AI followed by re-training of the “student” \cite{yosinski2014transferable}, \cite{chen2015net2net}. Alternatives to AI salvaging include model compression \cite{Bucila:2006}, knowledge {\it distillation} \cite{Hinton:2015}, and {\it privileged information} \cite{vapnik2017knowledge}. These approaches demonstrated substantial success in improving generalization capabilities of AIs as well as in reducing computational overheads \cite{SqueezeNet:2016}, in cases of knowledge transfer from larger AI to the smaller one. Notwithstanding, however, which of the above strategies is followed, their implementation often requires either significant resources including large training sets and power needed for training, or access to privileged information that may not necessarily be available to end-users. Thus new frameworks and approaches are needed.

In this contribution we provide new framework for automated, fast, and non-destructive process of knowledge spreading across AI systems of varying architectures. In this framework, knowledge transfer is accomplished by means of Knowledge Transfer Units comprising of mere linear functionals and/or their small cascades. Main mathematical ideas are rooted in measure concentration \cite{Gromov:1999}, \cite{GAFA:Gromov:2003}, \cite{Gibbs1902}, \cite{Levi1951}, \cite{Gorban:2007} and stochastic separation theorems \cite{GorbanTyukin:NN:2017} revealing peculiar properties of random sets in high dimensions.  We generalize some of the latter results here and show how these generalizations can be employed to build simple one-shot Knowledge Transfer algorithms between heterogeneous AI systems whose state may be represented by elements of linear vector space of sufficiently high dimension. Once knowledge has been transferred from one AI to another, the approach also allows  to ``unlearn'' new knowledge without the need to store a complete copy of the ``student'' AI is created prior to learning. We expect that the proposed framework may pave way for fully functional new phenomenon -- Nursery of AI systems in which AIs quickly learn from each other whilst keeping their pre-existing skills largely intact.

The paper is organized as follows. Section \ref{sec:background} contains mathematical background needed to justify the proposed knowledge transfer algorithms. In Section \ref{sec:results} we present two algorithms for transferring knowledge between a pair of AI systems in which one operates as a teacher and the other functions as a student. Section \ref{sec:examples} illustrates the approach with examples, and Section \ref{sec:conclusion} concludes the paper.

\section{Mathematical background}\label{sec:background}

%\subsection{$k$-tuples separation theorems}

Let the set
\[
\mathcal{M}=\{\boldsymbol{x}_1,\dots,\boldsymbol{x}_M\}
\]
be an i.i.d. sample from a distribution in $\Real^n$. Pick  another set
\[
\mathcal{Y}=\{\boldsymbol{x}_{M+1},\dots,\boldsymbol{x}_{M+k}\}
\]
from the same distribution at random. What is the probability that there is a linear functional separating $\mathcal{Y}$ from $\mathcal{M}$?

Below we provide three $k$-tuple separation theorems: for an equidistribution in $B_n(1)$ (Theorem \ref{theorem:k-tuples:ball} and \ref{theorem:k-tuples:ball:correlated}) and for a product probability measure with bounded support (Theorem \ref{theorem:k-tuples:cube}). These two special cases cover or, indeed, approximate broad range of practically relevant situations including e.g. Gaussian distributions (reduce asymptotically to equidistribution in $B_n(1)$ for $n$ large enough) and data vectors in which each attribute is a numerical and independent random variable.

%\subsubsection{Equidisrtibution in $B_n(1)$}

Consider the case when the underlying probability distribution is an equidistribution in the unit ball $B_n(1)$, and suppose that $\mathcal{M}=\{\boldsymbol{x}_1,\dots,\boldsymbol{x}_M\}$ and $\mathcal{Y}=\{\boldsymbol{x}_{M+1},\dots,\boldsymbol{x}_{M+k}\}$ are i.i.d. samples from this distribution. We are interested in determining the probability $\mathcal{P}_1(\mathcal{M},\mathcal{Y})$ that there exists a linear functional $l$ separating $\mathcal{M}$ and $\mathcal{Y}$. An estimate of this probability is provided in the following theorem

\begin{figure}
\centering
\begin{minipage}{0.45\textwidth}
\centering
\includegraphics[width=140pt]{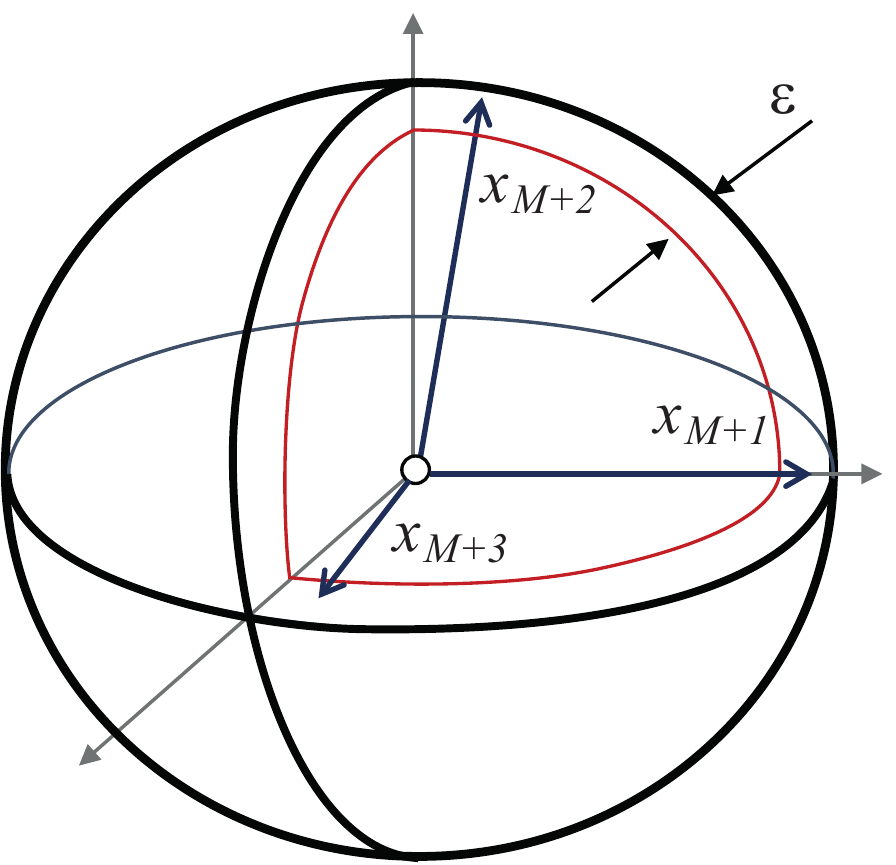}

(a)
\end{minipage}\hspace{5pt}
\begin{minipage}{0.45\textwidth}
\centering
\includegraphics[width=140pt]{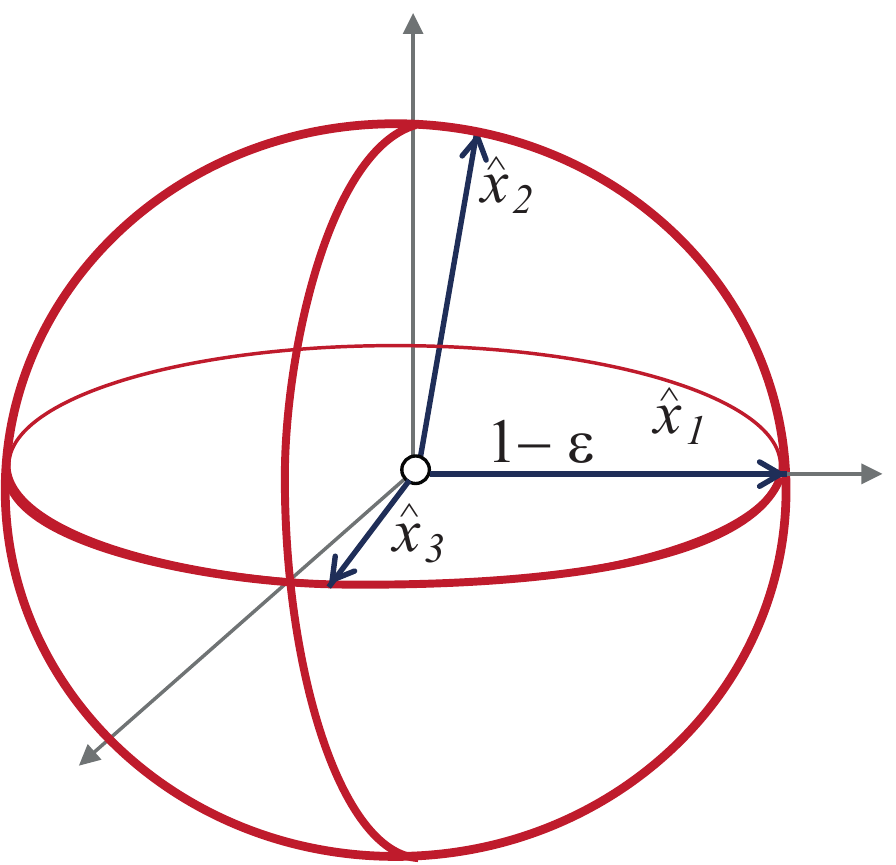}

(b)
\end{minipage}
\begin{minipage}{0.45\textwidth}
\centering
\includegraphics[width=110pt]{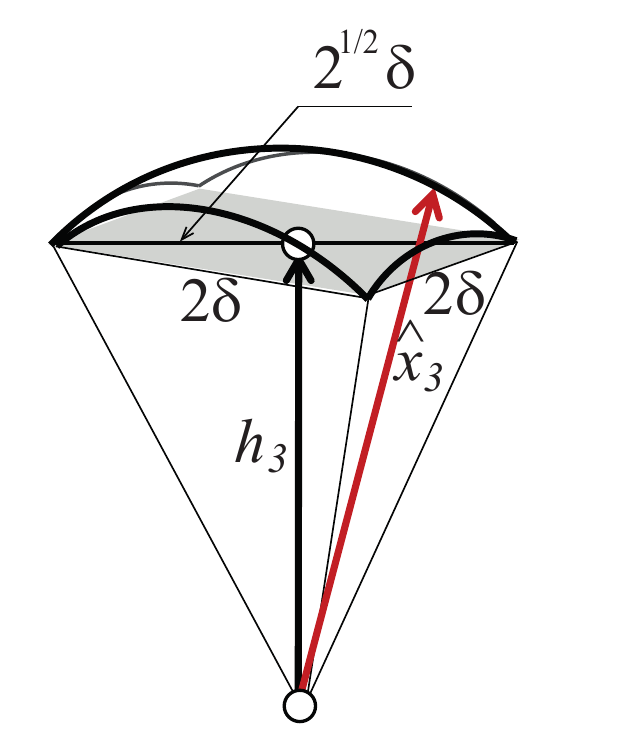}

(c)
\end{minipage}\hspace{5pt}
\begin{minipage}{0.45\textwidth}
\centering
\includegraphics[width=140pt]{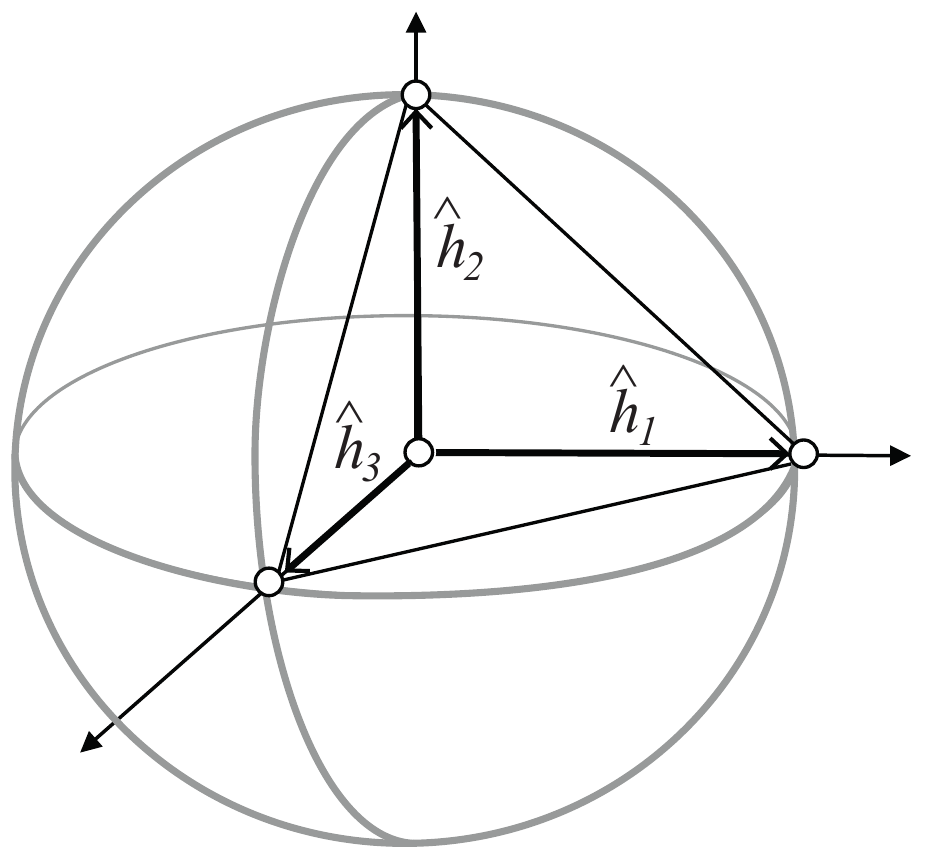}

(d)
\end{minipage}
\begin{minipage}{0.45\textwidth}
\centering
 \includegraphics[width=192pt]{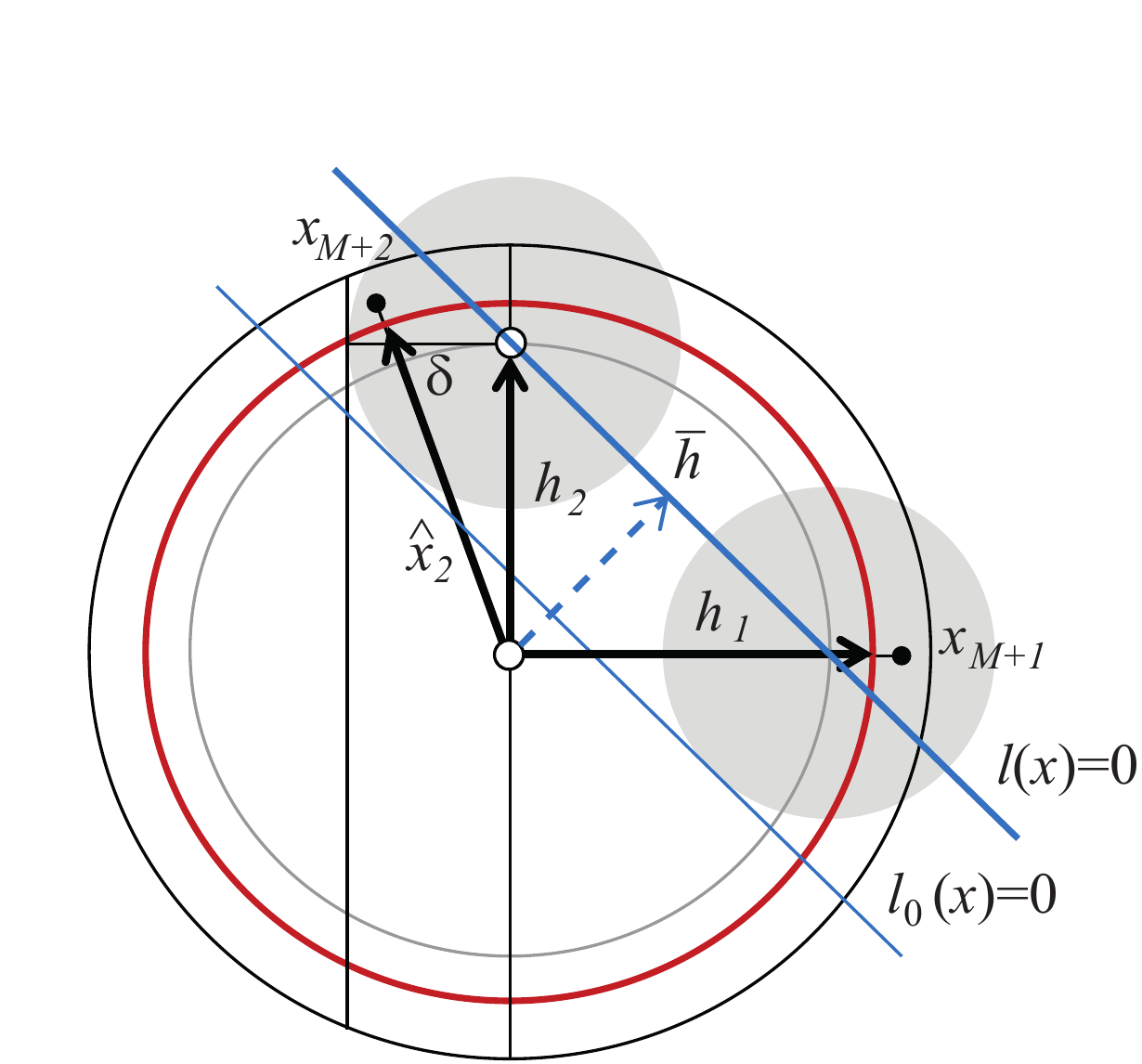}

(e)
\end{minipage}
\caption{Illustration to the proof of Theorem \ref{theorem:k-tuples:ball}. {\it Panel (a)} shows $\boldsymbol{x}_{M+1}$, $\boldsymbol{x}_{M+2}$ and $\boldsymbol{x}_{M+3}$ in the set $B_n(1)\setminus B_n(1-\varepsilon)$. {\it Panel (b)} shows $\hat{\boldsymbol{x}}_1$, $\hat{\boldsymbol{x}}_2$, and $\hat{\boldsymbol{x}}_3$ on the sphere $S_{n-1}(1-\varepsilon)$. {\it Panel (c)}: construction of $\boldsymbol{h}_3$. Note that $\|\boldsymbol{h}_3\|=\|\hat{\boldsymbol{x}}_3\|(1-2\delta^2)^{1/2}=(1-\varepsilon)(1-2\delta^2)^{1/2}$. {\it Panel (d)} shows simplex formed by orthogonal vectors $\hat{\boldsymbol{h}}_1,\hat{\boldsymbol{h}}_2,\hat{\boldsymbol{h}}_3$. {\it Panel (e)} illustrates derivation of functionals $l$ and $l_0$.}\label{fig:thm:ball}
\end{figure}

\begin{theorem}\label{theorem:k-tuples:ball} Let $\mathcal{M}=\{\boldsymbol{x}_1,\dots,\boldsymbol{x}_M\}$ and $\mathcal{Y}=\{\boldsymbol{x}_{M+1},\dots,\boldsymbol{x}_{M+k}\}$ be  i.i.d. samples from the equidisribution in $B_n(1)$. Then
\begin{equation}\label{eq:k-tuple_ball}
\begin{split}
{\mathcal{P}}_1(\mathcal{M},\mathcal{Y})& \geq \max_{\delta,\varepsilon} \ (1-(1-\varepsilon)^n)^{k} \prod_{m=1}^{k-1} \left(1-m \left(1-\delta^2\right)^{\frac{n}{2}}\right) \left(1 -
\frac{\Delta(\varepsilon,\delta,k)^\frac{n}{2}}{2}\right)^{M} \\
\Delta(\varepsilon,\delta,k)&= 1-\left[\frac{(1-\varepsilon)\sqrt{1-(k-1)\delta^2}}{\sqrt{k}}-(k-1)^{\frac{1}{2}}\delta\right]^2\\
& \mathrm{Subject} \ \mathrm{ to:}\\
& \delta, \varepsilon\in(0,1)\\
& 1-(k-1)\delta^2 \geq 0\\
& (k-1)(1-\delta^2)^{\frac{n}{2}}\leq 1\\
& \frac{(1-\varepsilon)\sqrt{1-(k-1)\delta^2}}{\sqrt{k}}-(k-1)^{\frac{1}{2}}\delta \geq 0.
\end{split}
\end{equation}
\end{theorem}
{\it Proof of Theorem \ref{theorem:k-tuples:ball}}. Given that elements in the set $\mathcal{Y}$ are independent, the probability $p_1$ that $\mathcal{Y} \subset B_n(1)\setminus B_n(1-\varepsilon)$ is
\[
p_1=(1-(1-\varepsilon)^n)^k.
\]
Consider an auxiliary set
\[
\hat{\mathcal{Y}}=\left\{\hat{\boldsymbol{x}}_{i}\in\Real^n \ | \ \hat{\boldsymbol{x}}_i=(1-\varepsilon)\frac{\boldsymbol{x}_{M+i}}{\|\boldsymbol{x}_{M+i}\|}, \ i=1,\dots,k \right\}.
\]
Vectors $\hat{\boldsymbol{x}}_i\in\hat{\mathcal{Y}}$ belong to the sphere of radius $1-\varepsilon$ centred at the origin (see Figure \ref{fig:thm:ball}, (b)).
According to \cite{GorTyu:2016} (proof of Proposition 3 and estimate (26)), the probability $p_2$ that for a given a given $\delta\in(0,1)$ all elements of  $\hat{\mathcal{Y}}$ are pair-wise $\delta/(1-\varepsilon)$-orthogonal, i.e.
\begin{equation}\label{eq:near-orthogonality}
%\left|\left\langle \frac{\hat{\boldsymbol{x}}_{i}}{\|\hat{\boldsymbol{x}}_{i}\|}, \frac{\hat{\boldsymbol{x}}_{j}}{\|\hat{\boldsymbol{x}}_{j}\|} \right\rangle %\right| \leq \delta \ \mbox{for all} \ i,j\in\{1,\dots,k\}, \ i\neq j,
\left|\cos\left(\hat{\boldsymbol{x}}_{i},\hat{\boldsymbol{x}}_{j} \right)\rangle\right| \leq \frac{\delta}{1-\varepsilon} \ \mbox{for all} \ i,j\in\{1,\dots,k\}, \ i\neq j,
\end{equation}
can be estimated from below as:
\[
p_2 \geq p_1 \prod_{m=1}^{k-1} \left(1-m \left(1-\delta^2\right)^{\frac{n}{2}}\right) = (1-(1-\varepsilon)^n)^k \prod_{m=1}^{k-1} \left(1-m \left(1-\delta^2\right)^{\frac{n}{2}}\right).
\]
for  $(k-1)(1-\delta^2)^{\frac{n}{2}}\leq 1$. Suppose now that (\ref{eq:near-orthogonality}) holds true. Let $\delta$ be chosen so that $1-(k-1)\delta^2\geq 0$. If this is the case than there exists a set of $k$ pair-wise orthogonal vectors
\[
\mathcal{H}=\{\boldsymbol{h}_1,\boldsymbol{h}_2,\dots,\boldsymbol{h}_{k}\}, \ \langle\boldsymbol{h}_i,\boldsymbol{h}_j\rangle =0, \ i,j\in\{1,\dots,k\}, \ i\neq j,
\]
such that (Figure \ref{fig:thm:ball}, (c))
\begin{equation}\label{eq:distance}
\|\hat{\boldsymbol{x}}_i-\boldsymbol{h}_i\|\leq (i-1)^{\frac{1}{2}}\delta, \ \|\boldsymbol{h}_i\|=(1-\varepsilon)(1-(i-1)\delta^2)^{\frac{1}{2}}, \ \mbox{for all} \ i\in\{1,\dots,k\}.
\end{equation}
Finally, consider the set
\[
\hat{\mathcal{H}}=\left\{\hat{\boldsymbol{h}}_{i}\in\Real^n \ | \ \hat{\boldsymbol{h}}_i=(1-\varepsilon)(1-(k-1)\delta^2)^{\frac{1}{2}}\frac{\boldsymbol{h}_{i}}{\|\boldsymbol{h}_{i}\|}, \ i=1,\dots,k \right\}
\]
The set $\hat{\mathcal{H}}$ belongs to the sphere of radius $(1-(k-1)\delta^2)^{\frac{1}{2}}$, and its $k$ elements are vertices of the corresponding $k-1$-simplex in $\Real^n$ (Figure \ref{fig:thm:ball}, (d)).

Consider the functional:
\[
l(\boldsymbol{x})=\left\langle \frac{\bar{\boldsymbol{h}}}{\|\bar{\boldsymbol{h}}\|}, \boldsymbol{x} \right\rangle - \frac{(1-\varepsilon)\sqrt{1-(k-1)\delta^2}}{\sqrt{k}}, \ \bar{\boldsymbol{h}}=\frac{1}{k}\sum_{i=1}^{k} \hat{\boldsymbol{h}}_i.
\]
Recall that if $\boldsymbol{e}_1,\dots,\boldsymbol{e}_k$ are orthonormal vectors in $\Real^n$ then $\|\boldsymbol{e}_1 + \boldsymbol{e}_2 + \cdots + \boldsymbol{e}_k \|^2 = k$. Hence $\left\|\sum_{i=1}^{k} \hat{\boldsymbol{h}}_i \right\|=\sqrt{k}(1-\varepsilon)\sqrt{1-(k-1)\delta^2}$, and we can conclude that
$l(\hat{\boldsymbol{h}}_i)=0$ and $l(\boldsymbol{h}_i)\geq 0$ for all $i=1,\dots,k$. According to (\ref{eq:distance}), $\|\hat{\boldsymbol{x}}_i -\boldsymbol{h}_i\|\leq (k-1)^{\frac{1}{2}}\delta$ for all $i=1,\dots,k$. Therefore the functional
\begin{equation}\label{eq:functional_1}
l_0(\boldsymbol{x})=l(\boldsymbol{x})+(k-1)^{\frac{1}{2}}\delta= \left\langle \frac{\bar{\boldsymbol{h}}}{\|\bar{\boldsymbol{h}}\|}, \boldsymbol{x} \right\rangle - \left(\frac{(1-\varepsilon)\sqrt{1-(k-1)\delta^2}}{\sqrt{k}}-(k-1)^{\frac{1}{2}}\delta\right)
\end{equation}
satisfies the following condition: $l_0(\hat{\boldsymbol{x}}_i)\geq 0$ and $l_0({\boldsymbol{x}}_{M+i})\geq 0$  for all $i=1,\dots,k$. This is illustrated with Figure \ref{fig:thm:ball}, (e).

The functional $l_0$ partitions the unit ball $B_n(1)$ into the union of two disjoint sets: the spherical cap $\mathcal{C}$
\begin{equation}\label{eq:cap}
\mathcal{C}=\{\boldsymbol{x}\in B_n(1) \ | l_0(\boldsymbol{x})\geq 0 \}
\end{equation}
and its complement in $B_n(1)$, $B_n(1)\setminus\mathcal{C}$. The volume $\mathcal{V}$ of the cap $\mathcal{C}$ can be estimated from above as
\[
\begin{split}
\mathcal{V}(\mathcal{C})&\leq \frac{\Delta(\varepsilon,\delta,k)^\frac{n}{2}}{2}, \\
\Delta(\varepsilon,\delta,k)&=1-\left[\frac{(1-\varepsilon)\sqrt{1-(k-1)\delta^2}}{\sqrt{k}}-(k-1)^{\frac{1}{2}}\delta\right]^2.
\end{split}
\]
Hence the probability $p_3$ that $l_0({\boldsymbol{x}}_i)<0$ for all $\boldsymbol{x}_i\in\mathcal{M}$ can be estimated from below as
\[
p_3\geq \left(1 - \frac{\Delta(\varepsilon,\delta,k)^\frac{n}{2}}{2}\right)^{M}.
\]
Therefore, for fixed $\varepsilon,\delta\in(0,1)$ chosen so that $\frac{(1-\varepsilon)\sqrt{1-(k-1)\delta^2}}{\sqrt{k}}-(k-1)^{\frac{1}{2}}\delta\geq 0$, the probability $p_4(\varepsilon,\delta)$ that $\mathcal{M}$ can be separated from $\mathcal{Y}$ by the functional $l_0$ can be estimated from below as:
\[
p_4(\varepsilon,\delta) \geq (1-(1-\varepsilon)^n)^{k} \prod_{m=1}^{k-1} \left(1-m \left(1-\delta^2\right)^{\frac{n}{2}}\right) \left(1 - \frac{\Delta(\varepsilon,\delta,k)^\frac{n}{2}}{2}\right)^{M}.
\]
Given that this estimate holds for all feasible values of $\varepsilon,\delta$, statement (\ref{eq:k-tuple_ball}) follows. $\square$

\begin{figure}
\centering
\includegraphics[width=200pt]{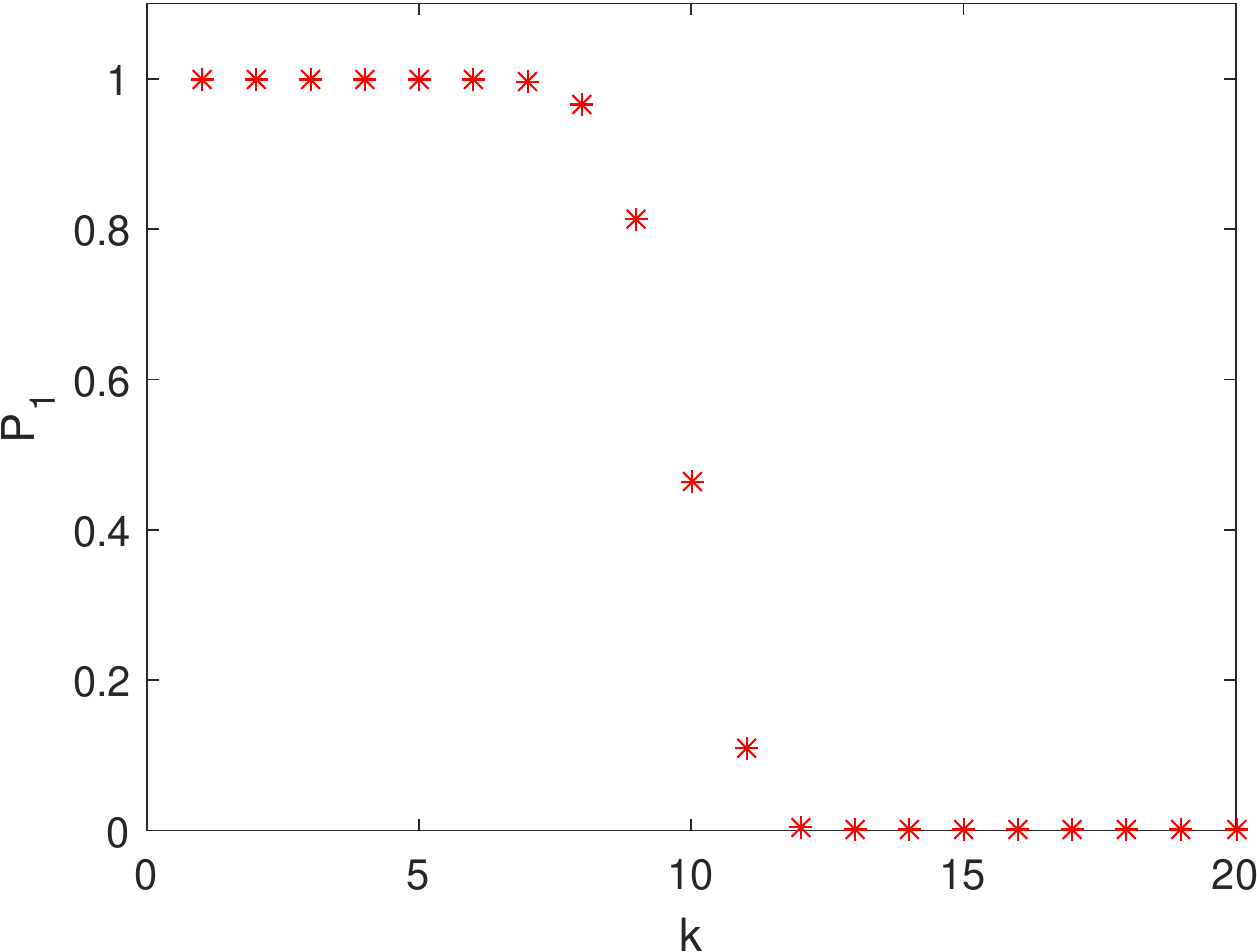}
\caption{Estimate (\ref{eq:k-tuple_ball}) of $\mathcal{P}_{1}(\mathcal{M},\mathcal{Y})$ as a function of $k$ for $n=2000$ and $M=10^5$.}\label{fig:thm:ball:probability}
\end{figure}

Figure \ref{fig:thm:ball:probability} shows how estimate (\ref{eq:k-tuple_ball}) of the probability $\mathcal{P}_{1}(\mathcal{M},\mathcal{Y})$ behaves, as a function of $|\mathcal{Y}|$ for fixed $M$ and $n$. As one can see from this figure, when $k$ exceeds some critical value ($k=9$ in this specific case), the lower bound estimate (\ref{eq:k-tuple_ball}) of the probability $\mathcal{P}_{1}(\mathcal{M},\mathcal{Y})$ drops. This is not surprising since the bound (\ref{eq:k-tuple_ball}) is a) based on rough, $L_\infty$-like, estimates, and b) these estimates are derived for just one class of separating functionals $l_0(\boldsymbol{x})$. Furthermore, no prior pre-processing and/or clustering was assumed for the $\mathcal{Y}$. An alternative estimate that allows us to account for possible clustering in the set $\mathcal{Y}$ is presented in Theorem \ref{theorem:k-tuples:ball:correlated}.

\begin{theorem}\label{theorem:k-tuples:ball:correlated} Let $\mathcal{M}=\{\boldsymbol{x}_1,\dots,\boldsymbol{x}_M\}$ and $\mathcal{Y}=\{\boldsymbol{x}_{M+1},\dots,\boldsymbol{x}_{M+k}\}$  be  i.i.d. samples from the equidistribution in $B_n(1)$. Let $\mathcal{Y}_c=\{\boldsymbol{x}_{M+r_1},\dots,\boldsymbol{x}_{M+r_m}\}$ be a subset of $m$ elements from $\mathcal{Y}$ such that
\begin{equation}\label{eq:k-tuples:assumption}
\beta_2 (m-1) \leq \sum_{r_j,\  r_j\neq r_i} \langle \boldsymbol{x}_{M+r_i}, \boldsymbol{x}_{M+r_j}\rangle \leq \beta_1 (m-1) \ \mbox{for all} \ i=1,\dots,m.
\end{equation}
Then
\begin{equation}\label{eq:k-tuple_ball:correlated}
\begin{split}
{\mathcal{P}}_1(\mathcal{M},\mathcal{Y}_c)& \geq \max_{\varepsilon\in(0,1)} (1-(1-\varepsilon)^n)^k \left(1 -
\frac{\Delta(\varepsilon,m)^\frac{n}{2}}{2}\right)^{M} \\
\Delta(\varepsilon,m)&=1- \frac{1}{m}\left(\frac{(1-\varepsilon)^2 + \beta_2 (m-1)}{\sqrt{1+(m-1)\beta_1}}\right)^2\\
& \mathrm{Subject} \ \mathrm{ to:}\\
& (1-\varepsilon)^2 + \beta_2 (m-1) > 0 \\
& 1+(m-1)\beta_1 >0.
\end{split}
\end{equation}
\end{theorem}
{\it Proof of Theorem \ref{theorem:k-tuples:ball:correlated}}. Consider the set $\mathcal{Y}$. Observe that $\|\boldsymbol{x}_{M+i}\|\geq 1-\varepsilon$, $\varepsilon\in(0,1)$, for all $i=1,\dots,k$, with probability $p_1= (1-(1-\varepsilon)^n)^k$. Consider now the vector  $\bar{\boldsymbol{y}}$
\[
\bar{\boldsymbol{y}}=\frac{1}{m}\sum_{i=1}^{m} \boldsymbol{x}_{M+r_i},
\]
and evaluate the following inner products
\[
\left\langle \frac{\bar{\boldsymbol{y}}}{\|\bar{\boldsymbol{y}}\|}, \boldsymbol{x}_{M+r_i} \right\rangle=\frac{1}{m \|\bar{\boldsymbol{y}}\|} \left(\langle \boldsymbol{x}_{M+r_i},\boldsymbol{x}_{M+r_i} \rangle +  \sum_{r_j, \ j\neq i} \langle \boldsymbol{x}_{M+r_i},\boldsymbol{x}_{M+r_j} \rangle\right), \ i=1,\dots,m.
\]
According to assumption (\ref{eq:k-tuples:assumption}), with probability $p_1$,
\[
\left\langle \frac{\bar{\boldsymbol{y}}}{\|\bar{\boldsymbol{y}}\|}, \boldsymbol{x}_{M+r_i} \right\rangle \geq \frac{1}{m \|\bar{\boldsymbol{y}}\|} \left((1-\varepsilon)^2 + \beta_2 (m-1) \right)
\]
and, respectively,
\[
\frac{1}{m}\left(1+(m-1)\beta_1 \right)\geq \langle\bar{\boldsymbol{y}},\bar{\boldsymbol{y}}\rangle\geq \frac{1}{m}\left((1-\varepsilon)^2 + \beta_2 (m-1)\right)
\]
Let $(1-\varepsilon)^2  + \beta_2(m-1) > 0$ and $(1-\varepsilon)^2  + \beta_1(m-1) > 0$. Consider the functional
\begin{equation}\label{eq:functional_2}
l_0(\boldsymbol{x})=\left\langle \frac{\bar{\boldsymbol{y}}}{\|\bar{\boldsymbol{y}}\|}, \boldsymbol{x} \right\rangle -  \frac{1}{\sqrt{m}}\left(\frac{(1-\varepsilon)^2 + \beta_2 (m-1)}{\sqrt{1+(m-1)\beta_1}}\right).
\end{equation}
It is clear that $l_0(\boldsymbol{x}_{M+r_i})\geq 0$ for all $i=1,\dots,m$ by the way the functional is constructed. The functional $l_0(\boldsymbol{x})$ partitions the ball $B_n(1)$ into two sets: the set $\mathcal{C}$ defined as in (\ref{eq:cap}) and its complement, $B_n(1)\setminus \mathcal{C}$. The volume $\mathcal{V}$ of the set $\mathcal{C}$ is bounded from above as
\[
\mathcal{V}(\mathcal{C})\leq \frac{\Delta(\varepsilon,m)^{\frac{n}{2}}}{2}
\]
where
\[
\Delta(\varepsilon,m)=1- \frac{1}{m}\left(\frac{(1-\varepsilon)^2 + \beta_2 (m-1)}{\sqrt{1+\beta_1 (m-1)}}\right)^2.
\]
Estimate (\ref{eq:k-tuple_ball:correlated}) now follows. $\square$
\begin{figure}
\centering
\includegraphics[width=200pt]{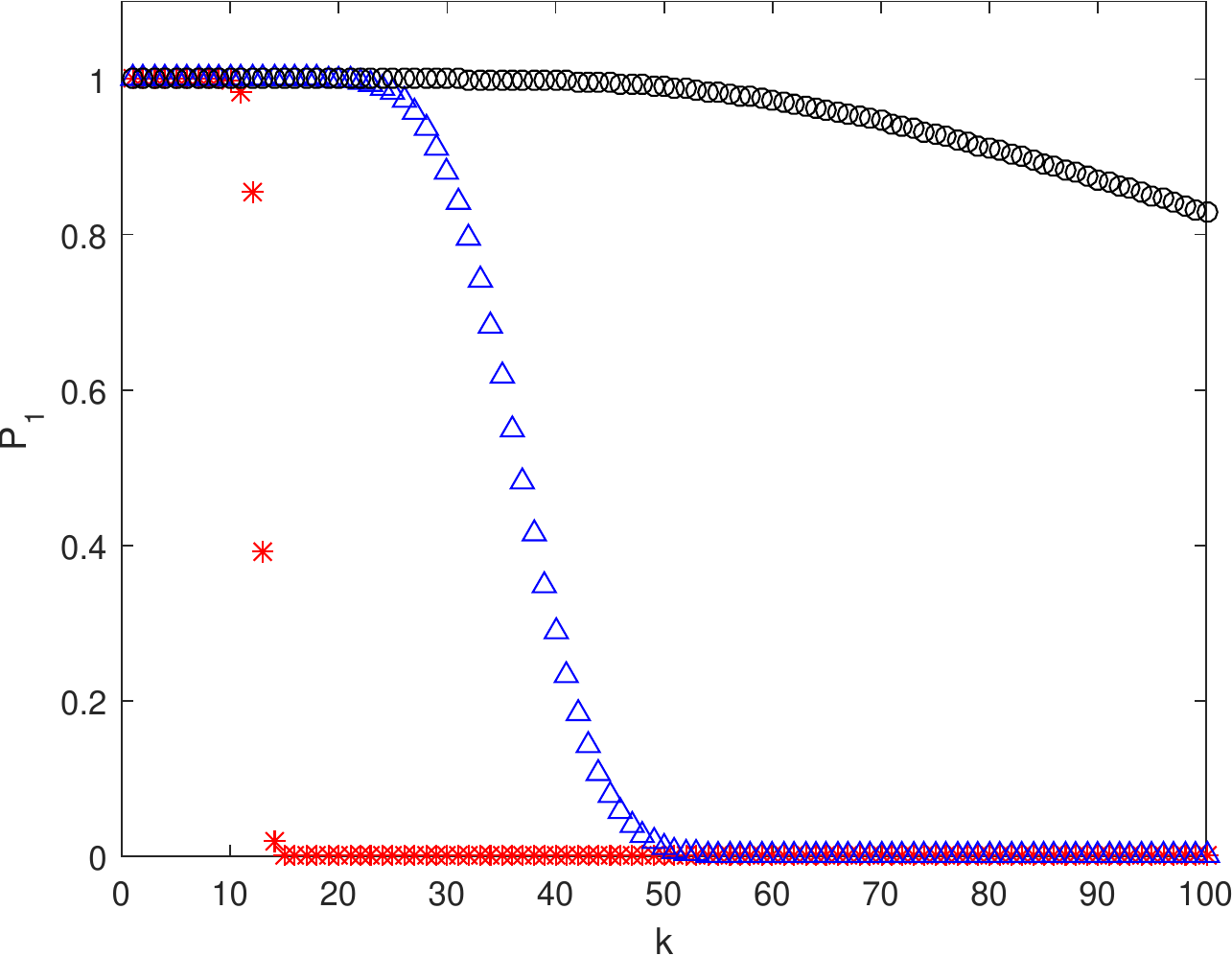}
\caption{Estimate (\ref{eq:k-tuple_ball:correlated}) of $\mathcal{P}_{1}(\mathcal{M},\mathcal{Y})$ as a function of $k$ for $n=2000$ and $M=10^5$. Red stars correspond to $\beta_1=0.5$, $\beta_2=0$. Blue triangles stand for $\beta_1=0.5$, $\beta_2=0.05$, and black circles stand for $\beta_1=0.5$, $\beta_2=0.07$.}\label{fig:thm:ball_correlated:probability}
\end{figure}

Examples of estimates (\ref{eq:k-tuple_ball:correlated}) for various parameter settings are shown in Fig. \ref{fig:thm:ball_correlated:probability}. As one can see, in absence of pair-wise strictly positive correlation assumption, $\beta_1=0$, the estimate's behavior, as a function of $k$, is similar to that of (\ref{eq:k-tuple_ball}). However, presence of moderate pair-wise positive correlation results in significant boosts to the values of $\mathcal{P}_1$.

\begin{rem}\normalfont Estimates (\ref{eq:k-tuple_ball}), (\ref{eq:k-tuple_ball:correlated}) for the probability $P_1(\mathcal{M},\mathcal{Y})$ that follow from Theorems \ref{theorem:k-tuples:ball}, \ref{theorem:k-tuples:ball:correlated} assume that the underlying probability distribution is an equidistribution in $B_n(1)$. They can, however, be generalized to equidistribuions in ellipsoids and Gaussian distributions (cf. \cite{GorTyuRom2016b}).
\end{rem}

 Note that proofs of Theorems \ref{theorem:k-tuples:ball}, \ref{theorem:k-tuples:ball:correlated} are constructive. Not only they provide estimates from below of the probability that two random i.i.d. drawn samples from $B_n(1)$ are linearly separable, but also they present the corresponding separating functionals explicitly as (\ref{eq:functional_1}) and (\ref{eq:functional_2}), respectively. The latter functionals are similar to Fisher linear discriminants. Whilst having explicit separation functionals is an obvious advantage from practical view point, the estimates that are associated with such functionals do not account for more flexible alternatives.  In what follows we present a generalization of the above results that accounts for such a possibility as well as extends applicability of the approach to samples from product distributions. The results are provided in Theorem \ref{theorem:k-tuples:cube}.

\begin{theorem}\label{theorem:k-tuples:cube} Consider the linear space $E=\mathrm{span} \{\boldsymbol{x}_j-\boldsymbol{x}_{M+1} \ | \ j=M+2,\dots, M+k\}$,  let the cardinality $|\mathcal{Y}|=k$ of the set $\mathcal{Y}$ be smaller than $n$. Consider the quotient space $\Real^n / E$. Let $Q(\boldsymbol{x})$ be a representation of $\boldsymbol{x}\in\Real^n$ in $\Real^n / E$, and let the coordinates of $Q(\boldsymbol{x}_{i})$, $i=1,\dots,M+1$  be independent random variables i.i.d. sampled from a product distribution in a unit cube with variances $\sigma_j>\sigma_0>0$, $1 \leq j\leq n-k+1$. Then for
\[
M\leq \frac{\vartheta}{3} \exp\left(\frac{(n-k+1)\sigma_0^4}{2}\right)-1
\]
with probability $p>1-\vartheta$ there is a linear functional separating $\mathcal{Y}$ and $\mathcal{M}$.
\end{theorem}
{\it Proof of Theorem \ref{theorem:k-tuples:cube}}.  Observe that, in the quotient space $\Real^n / E$, elements of the set
\[
\mathcal{Y}=\{\boldsymbol{x}_{M+1},\boldsymbol{x}_{M+1}+(\boldsymbol{x}_{M+2}-\boldsymbol{x}_{M+1}),\dots,\boldsymbol{x}_{M+1}+(\boldsymbol{x}_{M+k}-\boldsymbol{x}_{M+1})\}
\]
are vectors whose coordinates coincide with that of the quotient representation of $\boldsymbol{x}_{M+1}$. This means that the quotient representation of $\mathcal{Y}$ consists of a single element, $Q(\boldsymbol{x}_{M+1})$. Furthermore, dimension of $\Real^n/E$ is $n-k+1$.  Let $R_0^2=\sum_{i=1}^{n-k+1}\sigma_i^2$ and $\bar{Q}(\boldsymbol{x})=\mathbb{E}(Q(\boldsymbol{x}))$. According to Theorem 2 and Corollary 2 from \cite{GorbanTyukin:NN:2017}, for $\vartheta\in(0,1)$ and $M$ satisfying
\[
M\leq \frac{\vartheta}{3} \exp\left(\frac{(n-k+1)\sigma_0^4}{2}\right) - 1,
\]
with probability $p>1-\vartheta$  the following inequalities hold:
\[
\frac{1}{2} \leq \frac{\|Q(\boldsymbol{x}_{j})-\bar{Q}(\boldsymbol{x})\|^2}{R_0^2} \leq \frac{3}{2}, \  \left\langle \frac{Q(\boldsymbol{x}_{i})-\bar{Q}(\boldsymbol{x})}{R_0}, \frac{Q(\boldsymbol{x}_{M+1})-\bar{Q}(\boldsymbol{x})}{\|Q(\boldsymbol{x}_{M+1})-\bar{Q}(\boldsymbol{x})\|}  \right\rangle < \frac{1}{\sqrt{2}}
\]
for all $i,j$, $i\neq M+1$. This implies that the functional
\[
\ell_0(\boldsymbol{x})= \left\langle \frac{Q(\boldsymbol{x})-\bar{Q}(\boldsymbol{x})}{R_0}, \frac{Q(\boldsymbol{x}_{M+1})-\bar{Q}(\boldsymbol{x})}{\|Q(\boldsymbol{x}_{M+1})-\bar{Q}(\boldsymbol{x})\|}  \right\rangle - \frac{1}{\sqrt{2}}
\]
separates $\mathcal{M}$ and $\mathcal{Y}$ with probability $p > 1-\vartheta$. $\square$

\section{AI Knowledge Transfer Framework}\label{sec:results}

In this section we show how Theorems  \ref{theorem:k-tuples:ball}, \ref{theorem:k-tuples:ball:correlated} and \ref{theorem:k-tuples:cube} can be applied for developing a novel one-shot AI knowledge transfer framework. We will focus on the case of transfer knowledge between two AI systems, a teacher AI and a student AI, in which input-output behaviour of the student AI is evaluated by the teacher AI. In this setting,  assignment of AI roles, i.e. student or teaching, is beyond the scope of this manuscript. The roles are supposed to be pre-determined or otherwise chosen arbitrarily.

\subsection{General setup}

Consider two AI systems, a student AI, denoted as $\mathrm{AI}_s$, and a teacher AI, demoted as $\mathrm{AI}_t$. These legacy AI systems  process some {\it input} signals, produce {\it internal} representations of the input and return some {\it outputs}. We further assume that some {\it relevant} information about the input, internal signals, and outputs of $\mathrm{AI}_s$ can be combined into a common object, $\boldsymbol{x}$, representing, but not necessarily defining, the {\it state} of $\mathrm{AI}_s$. The objects $\boldsymbol{x}$ are assumed to be elements of $\Real^n$.

Over a period of activity system $\mathrm{AI}_s$ generates a set $\mathcal{S}$ of objects $\boldsymbol{x}$. Exact composition of the set $\mathcal{S}$ could  depend on a task at hand. For example, if $\mathrm{AI}_s$ is an image classifier, we may be interested only in a particular subset of $\mathrm{AI}_s$ input-output data related to images of a certain known class. Relevant inputs and outputs of $\mathrm{AI}_s$ corresponding to objects in $\mathcal{S}$ are then evaluated by the teacher, $\mathrm{AI}_t$. If $\mathrm{AI}_s$ outputs differ to that of $\mathrm{AI}_t$ for the same input then an error is registered in the system. Objects $\boldsymbol{x}\in\mathcal{S}$ associated with errors are combined into the set $\mathcal{Y}$. The procedure gives rise to two disjoint sets:
 \[
 \mathcal{M}=\mathcal{S}\setminus \mathcal{Y}, \ \mathcal{M}=\{\boldsymbol{x}_1,\dots,\boldsymbol{x}_{M}\}
 \]
and
 \[
 \mathcal{Y}=\{\boldsymbol{x}_{M+1},\dots,\boldsymbol{x}_{M+k}\}.
 \]
\begin{figure}
\centering
\includegraphics[width=\textwidth]{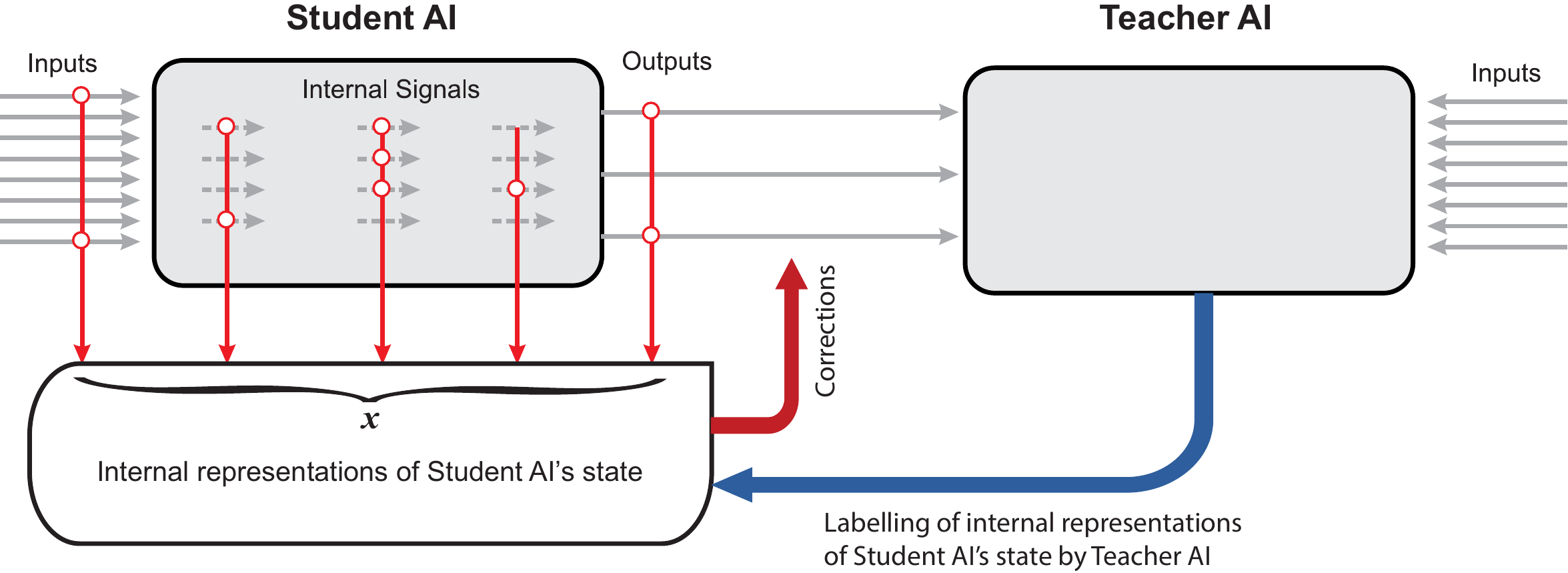}
\caption{AI Knowledge transfer diagram. $AI_s$ produces a set of its state representations, $\mathcal{S}$. The representations are labelled by $AI_t$ into the set of correct responses, $\mathcal{M}$, and the set of errors, $\mathcal{Y}$. The student system, $AI_s$, is then augmented by an additional ``corrector''  eliminating these errors.}\label{fig:knowledge_transfer_general}
\end{figure}
A diagram schematically representing the process is shown in Fig. \ref{fig:knowledge_transfer_general}. The knowledge transfer task is to ``teach'' $\mathrm{AI}_s$ so that with
\begin{itemize}
\item[ a)] $\mathrm{AI}_s$ does not make such errors
\item[ b)] existing competencies of $\mathrm{AI}_s$ on the set of inputs corresponding to internal states $\boldsymbol{x}\in\mathcal{M}$ are retained, and
\item[ c)] knowledge transfer from $\mathrm{AI}_t$ to $\mathrm{AI}_s$ is reversible in the sense that $\mathrm{AI}_s$ can ``unlearn'' new knowledge by modifying just a fraction of its parameters, if required.
\end{itemize}
Two algorithms for achieving such transfer knowledge are provided below.

\subsection{Knowledge Transfer Algorithms}

Our first algorithm, Algorithm \ref{alg:one_stage}, considers cases when {\it Auxiliary Knowledge Transfer Units}, i.e. functional additions to existing student $\mathrm{AI}_s$, are single linear functionals. The second algorithm, Algorithm \ref{alg:two_stages}, extends Auxiliary Knowledge Transfer Units to  two-layer cascades of linear functionals.

\begin{algorithm}
\caption{Single-functional AI Knowledge Transfer}\label{alg:one_stage}
\small
\begin{enumerate}
  \item \textbf{Pre-processing}
  \begin{enumerate}
    \item \textit{Centering}. For the given set $\mathcal{S}$, determine the set average, $\bar{\boldsymbol{x}}(\mathcal{S})$, and generate sets $\mathcal{S}_c$
        \[
          \begin{array}{ll}
        {\mathcal{S}_c}&=\{\boldsymbol{x}\in\Real^n \ | \boldsymbol{x}=\boldsymbol{\xi}-\bar{\boldsymbol{x}}(\mathcal{S}), \ \boldsymbol{\xi}\in\mathcal{S}\}, \\
        %{\mathcal{M}_c}&=\{\boldsymbol{x}\in\Real^n \ | \boldsymbol{x}=\boldsymbol{\xi}-\bar{\boldsymbol{x}}(\mathcal{S}), \ %\boldsymbol{\xi}\in\mathcal{M}\},\\
         {\mathcal{Y}_c}&=\{\boldsymbol{x}\in\Real^n \ | \boldsymbol{x}=\boldsymbol{\xi}-\bar{\boldsymbol{x}}(\mathcal{S}), \ \boldsymbol{\xi}\in\mathcal{Y}\}.
         \end{array}
        \]
    \item \textit{Regularization}. Determine covariance matrices $\Cov(\mathcal{S}_c)$, $\Cov(\mathcal{S}_c\setminus\mathcal{Y}_c)$ of the sets $\mathcal{S}_c$ and $\mathcal{S}_c\setminus\mathcal{Y}_c$. Let $\lambda_i(\Cov(\mathcal{S}_c))$, $\lambda_i(\Cov(\mathcal{S}_c\setminus\mathcal{Y}_c))$ be their corresponding eigenvalues, and $h_1, \dots, h_n$ be the eigenvectors of $\Cov(\mathcal{S}_c)$. If some of $\lambda_i(\Cov(\mathcal{S}_c))$, $\lambda_i(\Cov(\mathcal{S}_c\setminus\mathcal{Y}_c))$ are zero or if the ratio
       $ \frac{\max_i \{\lambda_i(\Sigma(\mathcal{S}_c))\}}{\min_i \{\lambda_i(\Sigma(S_c))\}}$
       is too large, project $\mathcal{S}_c$ and $\mathcal{Y}_c$  onto appropriately chosen set of $m<n$ eigenvectors, $h_{n-m+1},\dots,h_n$:
        \[
        \begin{array}{ll}
        {\mathcal{S}_r}&=\{\boldsymbol{x}\in\Real^n \ | \boldsymbol{x}=H^T \boldsymbol{\xi}, \ \boldsymbol{\xi}\in\mathcal{S}_c\}, \\
       % {\mathcal{M}_r}&=\{\boldsymbol{x}\in\Real^n \ | \boldsymbol{x}=H^T \boldsymbol{\xi}, \ \boldsymbol{\xi}\in\mathcal{M}_c\},\\
        {\mathcal{Y}_r}&=\{\boldsymbol{x}\in\Real^n \ | \boldsymbol{x}=H^T \boldsymbol{\xi}, \ \boldsymbol{\xi}\in\mathcal{Y}_c\},
        \end{array}
        \]
    where $H=\left(h_{n-m+1} \cdots h_n\right)$ is the matrix comprising of $m$ significant principal components of $\mathcal{S}_c$.
    \item \textit{Whitening}. For the centered and regularized dataset $\mathcal{S}_r$, derive its covariance matrix, $\Cov(\mathcal{S}_r)$, and generate whitened sets
  \[
  \begin{array}{ll}
  {\mathcal{S}_w}&=\{\boldsymbol{x}\in\Real^m \ | \boldsymbol{x}=\Cov(\mathcal{S}_r)^{-\frac{1}{2}} \boldsymbol{\xi}, \ \boldsymbol{\xi}\in\mathcal{S}_r\},\\
  %{\mathcal{M}_w}&=\{\boldsymbol{x}\in\Real^m \ | \boldsymbol{x}=\Cov(\mathcal{S}_r)^{-\frac{1}{2}} \boldsymbol{\xi}, \ \boldsymbol{\xi}\in\mathcal{M}_r\},\\
  {\mathcal{Y}_w}&=\{\boldsymbol{x}\in\Real^m \ | \boldsymbol{x}=\Cov(\mathcal{S}_r)^{-\frac{1}{2}} \boldsymbol{\xi}, \ \boldsymbol{\xi}\in\mathcal{Y}_r\},
  \end{array}
  \]
    \end{enumerate}
   \item \textbf{Knowledge transfer}
   \begin{enumerate}
    \item \textit{Clustering}. Pick $p\geq 1$, $p\leq k$, $p\in\Natural$, and partition the set $\mathcal{Y}_w$ into $p$ clusters $\mathcal{Y}_{w,1},\dots \mathcal{Y}_{w,p}$ so that elements of these clusters are, on average, pairwise positively correlated. That is there are $\beta_{1} \geq \beta_{2} > 0$ such that:
        \[
         \beta_2(|\mathcal{Y}_{w,i}|-1)\leq \sum_{\xi\in \mathcal{Y}_{w,i}\setminus\{\boldsymbol{x}\} } \langle\boldsymbol{\xi},\boldsymbol{x}\rangle  \leq \beta_1(|\mathcal{Y}_{w,i}|-1) \ \mbox{for any} \ \boldsymbol{x}\in \mathcal{Y}_{w,i}
        \]
    \item \textit{Construction  of Auxiliary Knowledge Units}. For each cluster $\mathcal{Y}_{w,i}$, $i=1,\dots,p$, construct separating linear functionals $\ell_i$:
    \[
    \begin{array}{ll}
    \ell_i(\boldsymbol{x})&=\left\langle\frac{\boldsymbol{w}_i}{\|\boldsymbol{w}_i\|},\boldsymbol{x}\right\rangle - c_i,\\
    \boldsymbol{w}_i&=\left(\Cov(\mathcal{S}_w \setminus \mathcal{Y}_{w,i}) + \Cov(\mathcal{Y}_{w,i}) \right)^{-1} \left(\bar{\boldsymbol{x}}(\mathcal{Y}_{w,i}) - \bar{\boldsymbol{x}}(\mathcal{S}_w\setminus\mathcal{Y}_{w,i}) \right)
    \end{array}
    \]
    where $\bar{\boldsymbol{x}}(\mathcal{Y}_{w,i})$,  $\bar{\boldsymbol{x}}(\mathcal{S}_w\setminus\mathcal{Y}_{w,i})$   are the averages of  $\mathcal{Y}_{w,i}$ and $\mathcal{S}_w \setminus \mathcal{Y}_{w,i}$, respectively, and $c_i$ is chosen as
   $c_i=\min_{\boldsymbol{\xi}\in \mathcal{Y}_{w,i}} \left\langle \frac{\boldsymbol{w}_i}{\|\boldsymbol{w}_i\|},\boldsymbol{\xi}\right\rangle$.

    \item \textit{Integration}. Integrate Auxiliary Knowledge Units into decision-making pathways of $\mathrm{AI}_s$. If, for an $\boldsymbol{x}$ generated by an input to $\mathrm{AI}_s$, any of $\ell_i(\boldsymbol{x})\geq 0$ then report $\boldsymbol{x}$ accordingly (swap labels, report as an error etc.)
    \end{enumerate}
  \end{enumerate}
\end{algorithm}

The algorithms comprise of two general stages, pre-processing stage and knowledge transfer stage. The purpose of the pre-processing stage is to regularize and ``sphere'' the data. This operation brings the setup close to the one considered in statements of Theorems \ref{theorem:k-tuples:ball}, \ref{theorem:k-tuples:ball:correlated}. The knowledge transfer stage constructs Auxiliary Knowledge Transfer Units in a way that is very similar to the argument presenteed in the proofs of Theorems \ref{theorem:k-tuples:ball} and \ref{theorem:k-tuples:ball:correlated}. Indeed, if $|\mathcal{Y}_{w,i}|\ll |\mathcal{S}_w \setminus\mathcal{Y}_{w,i}|$ then the term $\left(\Cov(\mathcal{S}_w \setminus \mathcal{Y}_{w,i}) + \Cov(\mathcal{Y}_{w,i}) \right)^{-1}$ is close to identity matrix, and the functionals $\ell_i$ are good approximations of (\ref{eq:functional_2}). In this setting, one might expect that performance of the knowledge transfer stage would be also closely aligned with the corresponding estimates (\ref{eq:k-tuple_ball}), (\ref{eq:k-tuple_ball:correlated}).

\begin{rem}\normalfont Note that the regularization step in the pre-processing stage ensures that the matrix $\Cov(\mathcal{S}_w \setminus \mathcal{Y}_{w,i}) + \Cov(\mathcal{Y}_{w,i})$ is non-singular. Indeed, consider
\[
\begin{array}{ll}
&\Cov(\mathcal{S}_w \setminus \mathcal{Y}_{w,i})= \frac{1}{|\mathcal{S}_w \setminus \mathcal{Y}_{w,i}|} \sum_{\boldsymbol{x}\in \mathcal{S}_w \setminus \mathcal{Y}_{w,i}} (\boldsymbol{x} - \bar{\boldsymbol{x}}(\mathcal{S}_w \setminus \mathcal{Y}_{w,i})) (\boldsymbol{x} - \bar{\boldsymbol{x}}(\mathcal{S}_w \setminus \mathcal{Y}_{w,i}))^{T}\\
&= \frac{1}{|\mathcal{S}_w \setminus \mathcal{Y}_{w,i}|} \left( \sum_{\boldsymbol{x}\in \mathcal{S}_w \setminus \mathcal{Y}_{w}} (\boldsymbol{x} - \bar{\boldsymbol{x}}(\mathcal{S}_w \setminus \mathcal{Y}_{w,i})) (\boldsymbol{x} - \bar{\boldsymbol{x}}(\mathcal{S}_w \setminus \mathcal{Y}_{w,i}))^{T} \right. + \\
& \left.\sum_{\boldsymbol{x}\in \mathcal{Y}_w \setminus \mathcal{Y}_{w,i}} (\boldsymbol{x} - \bar{\boldsymbol{x}}(\mathcal{S}_w \setminus \mathcal{Y}_{w,i})) (\boldsymbol{x} - \bar{\boldsymbol{x}}(\mathcal{S}_w \setminus \mathcal{Y}_{w,i}))^{T} \right).
\end{array}
\]
Denoting $d=\bar{\boldsymbol{x}}(\mathcal{S}_w \setminus \mathcal{Y}_{w,i})-\bar{\boldsymbol{x}}(\mathcal{S}_w\setminus \mathcal{Y}_{w})$ and rearranging the sum below as
\[
\begin{array}{ll}
&\sum_{\boldsymbol{x}\in \mathcal{S}_w \setminus \mathcal{Y}_{w}} (\boldsymbol{x} - \bar{\boldsymbol{x}}(\mathcal{S}_w \setminus \mathcal{Y}_{w,i})) (\boldsymbol{x} - \bar{\boldsymbol{x}}(\mathcal{S}_w \setminus \mathcal{Y}_{w,i}))^{T}= \\
&\sum_{\boldsymbol{x}\in \mathcal{S}_w \setminus \mathcal{Y}_{w}} (\boldsymbol{x}-\bar{\boldsymbol{x}}(\mathcal{S}_w \setminus \mathcal{Y}_{w})+d) (\boldsymbol{x}-\bar{\boldsymbol{x}}(\mathcal{S}_w \setminus \mathcal{Y}_{w})+d)^{T}=\\
& \sum_{\boldsymbol{x}\in \mathcal{S}_w \setminus \mathcal{Y}_{w}} (\boldsymbol{x}-\bar{\boldsymbol{x}}(\mathcal{S}_w \setminus \mathcal{Y}_{w})) (\boldsymbol{x}-\bar{\boldsymbol{x}}(\mathcal{S}_w \setminus \mathcal{Y}_{w}))^{T} + \\
&2 d \sum_{\boldsymbol{x}\in \mathcal{S}_w \setminus \mathcal{Y}_{w}}  (\boldsymbol{x}-\bar{\boldsymbol{x}}(\mathcal{S}_w \setminus \mathcal{Y}_{w}))^{T} + |\boldsymbol{x}\in \mathcal{S}_w \setminus \mathcal{Y}_{w}| d d^{T}\\
&= \sum_{\boldsymbol{x}\in \mathcal{S}_w \setminus \mathcal{Y}_{w}} (\boldsymbol{x}-\bar{\boldsymbol{x}}(\mathcal{S}_w \setminus \mathcal{Y}_{w})) (\boldsymbol{x}-\bar{\boldsymbol{x}}(\mathcal{S}_w \setminus \mathcal{Y}_{w}))^{T} + |\boldsymbol{x}\in \mathcal{S}_w \setminus \mathcal{Y}_{w}| d d^{T}
\end{array}
\]
we obtain that $\Cov(\mathcal{S}_w \setminus \mathcal{Y}_{w,i})$ is non-singular as long as the sum $\sum_{\boldsymbol{x}\in \mathcal{S}_w \setminus \mathcal{Y}_{w}} (\boldsymbol{x}-\bar{\boldsymbol{x}}(\mathcal{S}_w \setminus \mathcal{Y}_{w})) (\boldsymbol{x}-\bar{\boldsymbol{x}}(\mathcal{S}_w \setminus \mathcal{Y}_{w}))^{T}$ is non-singular. The latter property, however, is guaranteed by the regularization step in Algorithm \ref{alg:one_stage}.
\end{rem}

\begin{rem}\normalfont Clustering at Step 2.a can be achieved by classical $k$-means algorithms \cite{Lloyd:1982} or any other method (see e.g. \cite{DudaHart}) that would group elements of $\mathcal{Y}_w$ into clusters according to spatial proximity.
\end{rem}

\begin{rem} \normalfont Auxiliary Knowledge Transfer Units in Step 2.b of Algorithm \ref{alg:one_stage} are derived in accordance with standard Fisher linear discriminant formalism. This, however, need not be the case, and other methods such as e.g. support vector machines \cite{Vapnik2000}  could be employed for this purpose there. It is worth mentioning, however, that support vector machines might be prone to overfitting  \cite{Han:2014} and their training often involves iterative procedures such as e.g. sequential quadratic minimization \cite{Platt:1998}.

Furthermore, instead of the sets $\mathcal{Y}_{w,i}$, $\mathcal{S}_w \setminus \mathcal{Y}_{w,i}$ one could use a somewhat more aggressive division: $\mathcal{Y}_{w,i}$ and $\mathcal{S}_w \setminus \mathcal{Y}_{w}$, respectively.
\end{rem}

Depending on configuration of samples $\mathcal{S}$ and $\mathcal{Y}$,  Algorithm \ref{alg:one_stage} may occasionally create knowledge transfer units, $\ell_i$, that are ``filtering'' errors too aggressively. That is some $\boldsymbol{x}\in\mathcal{S}_w\setminus\mathcal{Y}_w$ may accidentally trigger non-negative response, $\ell_i(\boldsymbol{x})\geq 0$, and as a result of this their corresponding inputs to $\mathrm{A}_s$ could be ignored or mishandled. To mitigate this, one can increase the number of clusters and knowledge transfer units, respectively. This will increase the probability of successful separation and hence alleviate the issue. On the other hand, if increasing the number of knowledge transfer units is not desirable for some reason, then two-functional units could be a feasible remedy. Algorithm \ref{alg:two_stages} presents a procedure for such an improved AI Knowledge Transfer.

\begin{algorithm}
\caption{Two-functional AI Knowledge Transfer}\label{alg:two_stages}
\small

\begin{enumerate}
 \item \textbf{Pre-processing}. Do as in Step 1 in Algorithm \ref{alg:one_stage}
  \item \textbf{Knowledge Transfer}
 \begin{enumerate}
    \item \textit{Clustering}. Do as in Step 2.a in Algorithm \ref{alg:one_stage}
    \item \textit{Construction of Auxiliary Knowledge Units}.
    \begin{algorithmic}[1]
      \State Do as in Step 2.b in Algorithm \ref{alg:one_stage}. At the end of this step {\it first-stage} functionals $\ell_i$, $i=1,\dots,p$ will be derived.
      \State For each set $\mathcal{Y}_{w,i}$, $i=1,\dots,p$, evaluate the functionals $\ell_i$ for all $\boldsymbol{x}\in\mathcal{S}_w\setminus\mathcal{Y}_{w,i}$ and identify elements $\boldsymbol{x}$ such that $\ell_i(\boldsymbol{x})\geq 0$ and $\boldsymbol{x}\in \mathcal{S}_w\setminus\mathcal{Y}_w$ (incorrect error assignment). Let $\mathcal{Y}_{e,i}$ be the set containing such elements $\boldsymbol{x}$.
      \State \textbf{If} (there is an $i\in\{1,\dots,p\}$ such that $|\mathcal{Y}_{e,i}| +|\mathcal{Y}_{w,i}|>m$)
       \textbf{then} increment the value of $p$: $p\leftarrow p+1$, and return to Step 2.a.
       \State \textbf{If} (all sets $\mathcal{Y}_{e,i}$ are empty) \textbf{then} proceed to Step 2.c.
      \State For each pair of  $\ell_i$ and $\mathcal{Y}_{w,i}\cup \mathcal{Y}_{e,i}$ with $\mathcal{Y}_{e,i}$ not empty, project orthogonally sets $\mathcal{Y}_{w,i}$ and $\mathcal{Y}_{e,i}$ onto the hyperplane $\ell_i(\boldsymbol{x})=0$ and form the sets $\mathcal{L}_i(\mathcal{Y}_{w,i})$ and $\mathcal{L}_i(\mathcal{Y}_{e,i})$ :
      \[
      \begin{array}{ll}
      \mathcal{L}_i(\mathcal{Y}_{w,i})&=\left\{\boldsymbol{x}\in\Real^m \ | \ \boldsymbol{x}=\left(I_m - \frac{\boldsymbol{w}_i \boldsymbol{w}_i^T}{\|\boldsymbol{w}_i\|^2}\right)\boldsymbol{\xi} + \frac{c_i \boldsymbol{w}_i}{\|\boldsymbol{w}_i\|},  \ \boldsymbol{\xi}\in \mathcal{Y}_{w,i}\right\},\\
      \mathcal{L}_i(\mathcal{Y}_{e,i})&=\left\{\boldsymbol{x}\in\Real^m \ | \ \boldsymbol{x}=\left(I_m - \frac{\boldsymbol{w}_i \boldsymbol{w}_i^T}{\|\boldsymbol{w}_i\|^2}\right)\boldsymbol{\xi} + \frac{c_i \boldsymbol{w}_i}{\|\boldsymbol{w}_i\|}, \ \boldsymbol{\xi}\in \mathcal{Y}_{e,i}\right\}.
      \end{array}
      \]
      \State Construct a linear functional $\ell_{2,i}$ separating $\mathcal{L}_i(\mathcal{Y}_{w,i})$ from $\mathcal{L}_i(\mathcal{Y}_{e,i})$ so that $\ell_{2,i}(\boldsymbol{x})\geq 0$ for all $\boldsymbol{x}\in \mathcal{Y}_{w,i}$ and $\ell_{2,i}(\boldsymbol{x})< 0$ for all $\boldsymbol{x}\in \mathcal{Y}_{e,i}$.
          \end{algorithmic}
    \item \textit{Integration}. Integrate Auxiliary Knowledge Units into decision-making pathways of $\mathrm{AI}_s$. If, for an $\boldsymbol{x}$ generated by an input to $\mathrm{AI}_s$, any of the predicates $(\ell_i(\boldsymbol{x})\geq 0)\wedge (\ell_{2,i}(\boldsymbol{x})\geq 0)$ hold true then report $\boldsymbol{x}$ accordingly (swap labels, report as an error etc.).
    \end{enumerate}
\end{enumerate}
\end{algorithm}

In what follows we illustrate the approach as well as the application of the proposed  Knowledge Transfer algorithms in a relevant problem of a computer vision system design for pedestrian detection in live video streams.

\section{Example}\label{sec:examples}

Let $AI_s$ and $AI_t$ be two systems developed, e.g. for the purposes of pedestrian detection in live video streams. Technological progress in embedded systems and availability of platforms such as e.g. Nvidia Jetson TX2 made hadrware deployment of such AI systems at the edge of computer vision processing pipelines feasible. These AI systems, however, lack computational power  to run state-of-the-art large scale object detection solutions such as e.g. ResNet \cite{ResNet} in real-time. Here we demonstrate that to compensate for this lack of power, AI Knowledge Transfer can be successfully employed. In particular, we suggest that the edge-based system is ``taught'' by the state-of-the-art teacher in a non-iterative and near-real time way. Since our building blocks are linear functionals, such learning will not lead to significant computational overheads. At the same time, as we will show later, the proposed AI Knowledge Transfer will result in a major boost to the system's performance in the conditions of the experiment.

\subsection{Definition of $AI_s$ and $AI_t$ and rationale}

In our experiments, the teacher AI, $AI_t$, was modeled by a deep Convolutional Network,  ResNet 18 \cite{ResNet} with circa $11$M trainable parameters. The network was trained on a ``teacher'' dataset comprised of $5.2$M non-pedestrian (negatives), and $600$K  pedestrian (positives) images. The student AI, $AI_s$, was modelled by a linear classifier with HOG features \cite{Dalal:2005} and $2016$ trainable parameters. The values of these parameters were the result of $AI_s$ training on a ``student'' dataset, a sub-sample of the ``teacher'' dataset comprising of $55$K positives and $130$K negatives, respectively. This choice of $AI_s$ and $AI_t$ systems enabled us to emulate interaction between edge-based AIs and their more powerful counterparts that could be deployed on larger servers or computational clouds.

\begin{figure}
\centering
\includegraphics[width=300pt]{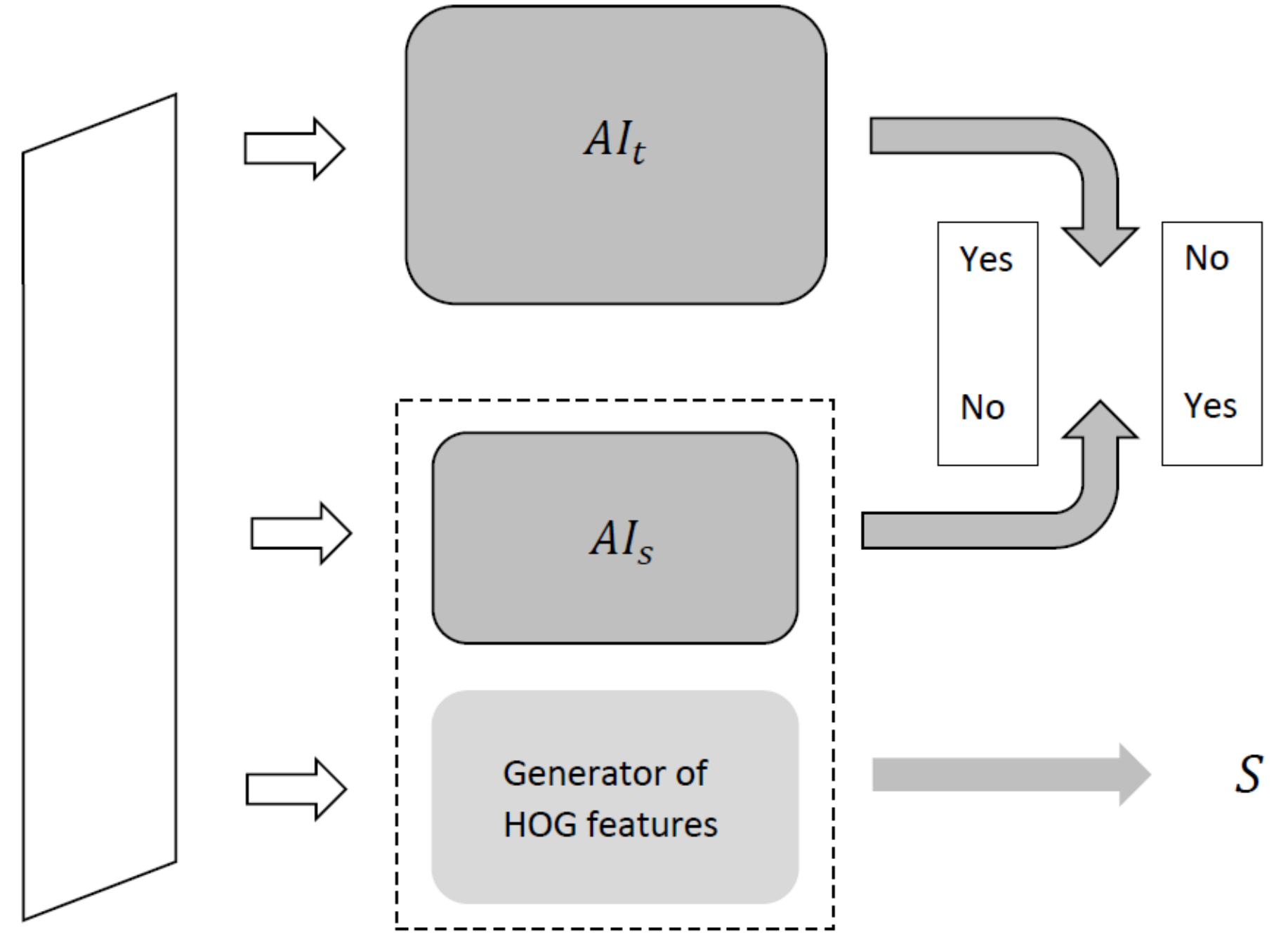}
\caption{Knowledge transfer diagram between ResNet and HOG-SVM object detectors}\label{fig:knowledge_transfer_example}
\end{figure}

Moreover, to make the experiment more realistic, we assumed that internal states of both systems are inaccessible for direct observation. To generate sets $\mathcal{S}$ and $\mathcal{Y}$ required in Algorithms \ref{alg:one_stage} and \ref{alg:two_stages} we augmented system $AI_s$ with an external generator of HOG features of the same dimension. We assumed, however, that covariance matrices of positives and negatives from the ``student'' dataset are available for the purposes of knowledge transfer. A diagram representing this setup is shown in Figure \ref{fig:knowledge_transfer_example}.
A candidate image is evaluated by two systems simultaneously as well as by a HOG features generator. The latter generates $2016$ dimensional vectors of HOGs and stores these vectors in the set $\mathcal{S}$. If outputs of $AI_s$ and $AI_t$ do not match the corresponding feature vector is added to the set $\mathcal{Y}$.

\subsection{Error types}

In this experiment we consider and address two types of errors: false positives (Type I errors) and false negatives (Type II errors). The error types were determined as follows. An error is deemed as {\it false positive} if $AI_s$ reported presence of a correctly sized  full-figure image of pedestrian in a given image patch whereas  no such object was there. Similarly, an error is deemed as {\it false negative} if a pedestrian was present in the given image patch but $AI_s$ did not report it there.

In our setting, evaluation of an image patch by $AI_t$ (ResNet) took $0.01$ sec on Nvidia K80 which was several orders slower than that of $AI_s$ (linear HOG-based classifier). Whilst such behavior was expected, this imposed technical limitations on the process of mitigating errors of Type II. Each frame from our testing video produced $400$K image patches to test. Evaluation of all these candidates by our chosen $AI_t$ is prohibitive computationally. To overcome this technical difficulty we tested only a limited subset of image proposals with regards to these error type. To get a computationally viable number of proposals for false negative testing, we increased sensitivity of the HOG-based classifier by lowering its detection threshold from $0$ to $-0.3$. This way our linear classifier with lowered threshold acted as a filter letting through more true positives at the expense of large number of false positives. In this operational mode, Knowledge Transfer Unit were tasked to separate true positives from negatives in accordance with object labels supplied by $AI_t$.

\subsection{Datasets}

The approach was tested on two benchmark videos: LINTHESCHER sequence \cite{Ess:2008} created by ETHZ and comprised of 1208 frames and NOTTINGHAM video \cite{Nottingham} containing 435 frames of live footage taken with an action camera. In what follows we will refer to these videos as ETHZ and NOTTINGHAM videos, respectively. ETHZ  video contains complete images of 8435 pedestrians, whereas NOTTINGHAM video has 4039 full-figure images of pedestrians.

%For testing this approach we used two videos. One of these videos was provided by Swiss Federal Institute of Technology in Zurich (ETHZ) and contains 1209 frames recorded on streets of Zurich[]. The second video was recorded by ourselves on streets of Nottingham[]. The second video contains 435 frames. We will refer to these videos as ETHZ and NOTTINGHAM videos respectively. For each of these videos we have created ground truth: almost every human full figure was marked with a bounding box. Ground truth detections can be used for benchmarking any type of detector. By having all possible detections at hand one can easily calculate true positive and false positive rates, which will be referred as TP and FP respectively. We also differentiate testing and evaluation stage. During the testing stage we perform detection obtaining and collecting all found detections. During the evaluation stage we actually calculate $TP$ and $FP$ rates.

%
%%\section{Datasets}\label{sec:datasets}
%Dataset we used for testing ideas described above is comprised of 1) the original dataset of pedestrian images(54000 samples) and non-pedestrian images(104000 images), 2) NOTTINGHAM video [] containing 3300 possible detections 3) LINTHESCHER video containing 6700 full figure images. Set 1) is used for training of HOG SVM classifier, whereas 2) and 3) are used for validation purposes. These sets do not have any common elements.
%
%ROC curves building

\subsection{Results}

\begin{figure}
\centering
\includegraphics[width=0.7\textwidth]{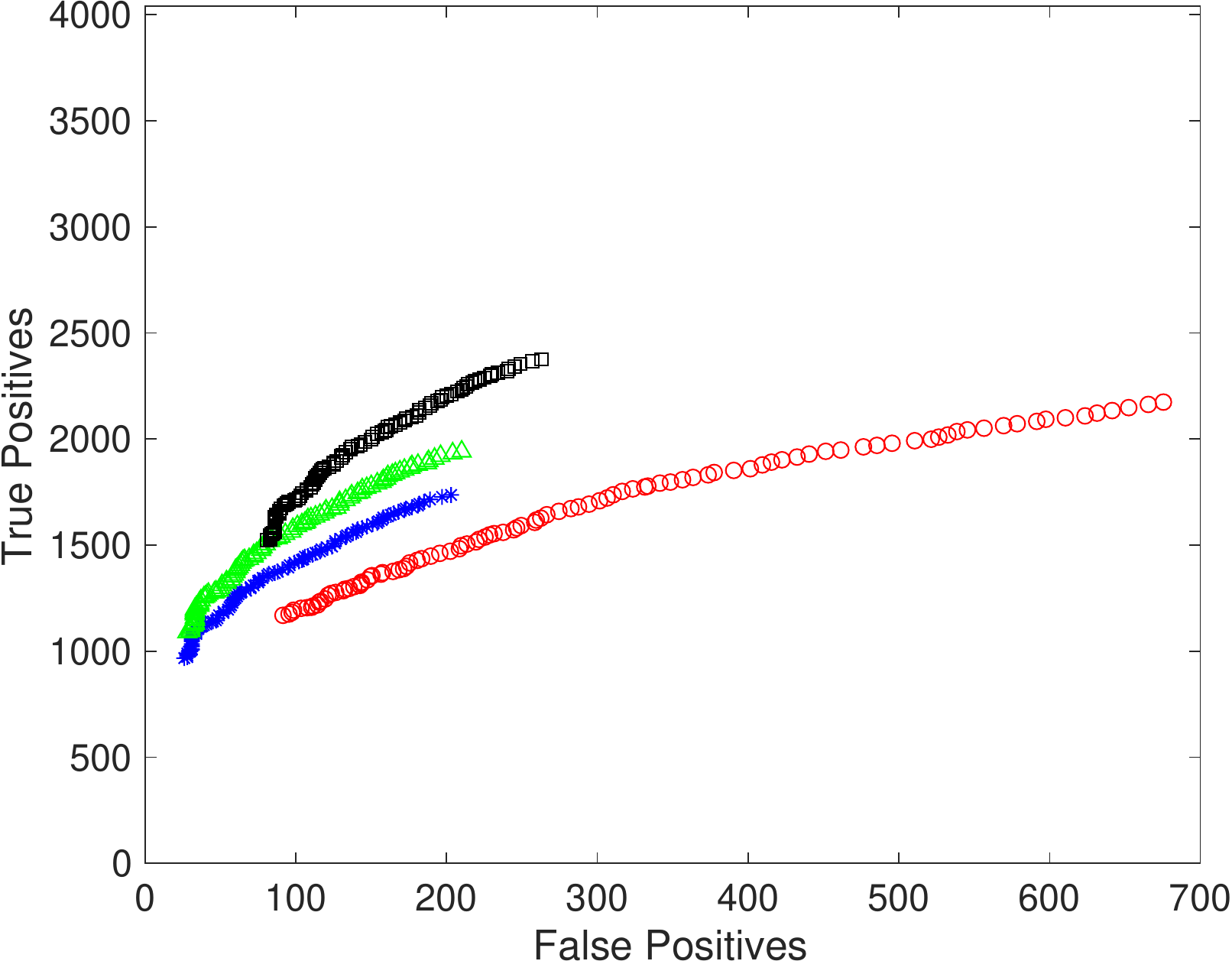}
\caption{True positives as a function of false positives for NOTTINGHAM video. }\label{fig:nottingham}
\end{figure}

\begin{figure}
\centering
\includegraphics[width=0.7\textwidth]{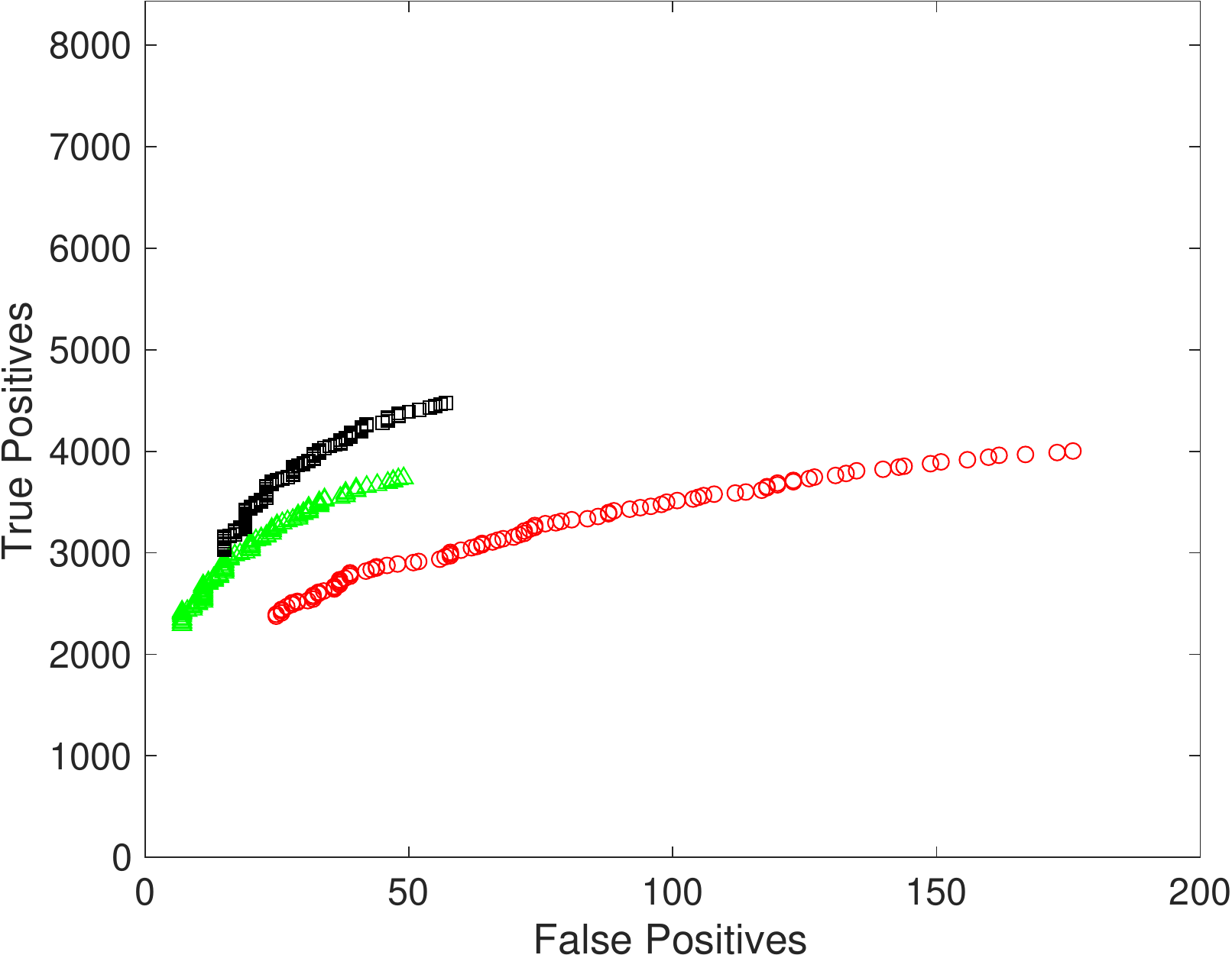}
\caption{True positives as a function of false positives for ETHZ video. }\label{fig:ethz}
\end{figure}

Performance and application of Algorithms \ref{alg:one_stage}, \ref{alg:two_stages} for NOTTINGHAM and ETHZ videos are summarized in Fig. \ref{fig:nottingham} and \ref{fig:ethz}. Each curves in these figures is produced by varying the values of decision-making threshold in the HOG-based linear classifier.  Red circles in Figure \ref{fig:nottingham} show true positives as  a function of false positives for the original linear classifier based on HOG features. Parameters of the classifier were set in accordance with Fisher linear discriminant formulae. Blue stars correspond to $AI_s$ after Algorithm \ref{alg:one_stage} was applied to mitigate errors of Type I in the system. The value of $p$ (number of clusters) in the algorithm was set to be equal to $5$. Green triangles illustrate application of Algorithm \ref{alg:two_stages} for the same error type. Here Algorithm \ref{alg:two_stages} was slightly modified so that the resulting Knowledge Transfer Unit had only one functional $\ell_2$. This was due to the low number of errors reaching stage two of the algorithm.  Black squares correspond to $AI_s$ after application of Algorithm \ref{alg:two_stages} (error Type I) followed by application of Algorithm \ref{alg:two_stages} to mitigate errors of Type II.

Figure \ref{fig:ethz} shows performance of the algorithms for ETHZ sequence. Red circles show performance of the original $AI_s$, green triangles correspond to $AI_s$ supplemented with Knowledge Transfer Units derived using Algorithm \ref{alg:two_stages} for errors of Type I. Black squares correspond to subsequent application of Algorithm \ref{alg:two_stages} dealing with errors of Type II.

\begin{figure}
\includegraphics[width=\textwidth]{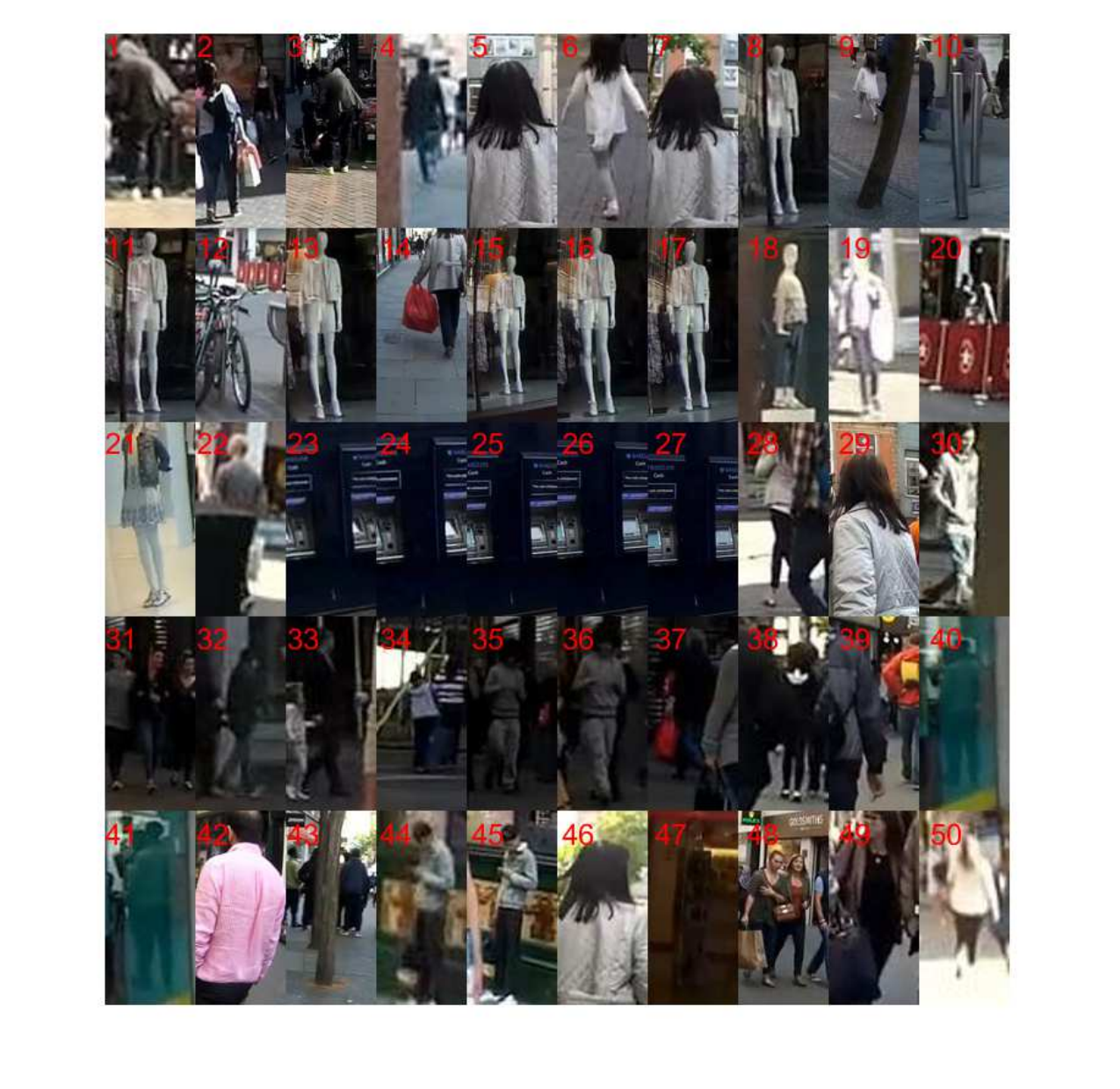}
\caption{False Positives induced by the teacher AI, $AI_t$.}\label{fig:errors}
\end{figure}

In all these cases, supplementing $AI_s$ with Knowledge Transfer Units constructed with the help of Algorithms \ref{alg:one_stage}, \ref{alg:two_stages} for both error types resulted in significant boost to $AI_s$ performance. Observe that in both cases application of Algorithm \ref{alg:two_stages} to address errors of Type II has led to noticeable increases of  numbers of false positives in the system at the beginning of the curves. Manual inspection of these false positives revealed that these  errors are exclusively due mistakes of $AI_t$ itself. For the sake of illustration, these errors for NOTTINGHAM video are shown in Fig. \ref{fig:errors}. These errors contain genuine false positives (images 12, 23-27) as well as mismatches by size (e.g. 1-7), and look-alikes (images 8,11,13,15-17).

\section{Conclusion}\label{sec:conclusion}

In this work we proposed a framework for instantaneous knowledge transfer between AI systems whose internal state used for decision-making can be described by elements of a high-dimensional vector space. The framework enables development of non-iterative algorithms for knowledge spreading between legacy AI systems with heterogeneous non-identical architectures and varying computing capabilities. Feasibility of the framework was illustrated with an example of knowledge transfer between two AI systems for automated pedestrian detection in video streams.

 In the basis of the proposed knowledge transfer framework are separation theorems (Theorem \ref{theorem:k-tuples:ball} -- \ref{theorem:k-tuples:cube}) stating peculiar properties of large but finite random samples in high dimension. According to these results, $k<n$ random i.i.d. elements can be separated form  $M\gg n$ randomly selected elements i.i.d. sampled from the same distribution by few linear functionals, with high probability. The theorems are proved for equidistributions in a ball and in a cube. The results can be trivially generalized to equidistributions in ellipsoids and Gaussian distributions. Generalizations to other meaningful distributions is the subject of our future work.

\section*{Acknowledgments}

The work was supported by Innovate UK Technology Strategy Board (Knowledge Transfer Partnership grants KTP009890 and KTP010522).

%\section*{References}

\bibliography{knowledge_transfer}

\end{document}